\documentclass{article}

\usepackage{spconf,graphicx}
\usepackage{amsmath,dsfont}
\usepackage{amssymb,latexsym}
\usepackage{amsfonts}
\usepackage{amscd}
\usepackage{algorithm}
\usepackage{algorithmic}
\usepackage{multirow}
\usepackage{changepage}
\usepackage{times}
\usepackage{float}
\usepackage{graphics,subfigure}
\usepackage[latin1]{inputenc}
\usepackage{graphicx}
\usepackage{color}
\usepackage{relsize,balance}
\usepackage[noadjust]{cite}

\def\cred{\textcolor{black}}

\newtheorem{assumption}{\bf{Assumption}}

\newtheorem{theorem}{\hskip\parindent\bf{Theorem}}[section]

\graphicspath{{figures/}}

\newcommand{\cb}[1]{{\boldsymbol{#1}}}
\newcommand{\cp}[1]{\ifmmode {\mathcal{#1}}\else ${\mathcal{#1}}$\fi}
\newcommand{\bs}{\boldsymbol}

\def\tr{\text{trace}}

\newcommand{\bI}{\boldsymbol{I}}

\newcommand{\T}{{\bs T}}
\newcommand{\bp}{{\bs p}}

\newcommand{\mG}{{\bs G}}
\newcommand{\mI}{{\bs I}}
\newcommand{\vc}{{\bs c}}

\newcommand{\bc}{\boldsymbol{c}}

\newcommand{\bv}{\boldsymbol{v}}
\newcommand{\bvD}{\boldsymbol{v}_\omega}
\newcommand{\bx}{\boldsymbol{x}}
\newcommand{\bxw}[1]{\bx_{\omega_{#1}}}

\newcommand{\by}{\boldsymbol{y}}

\newcommand{\bC}{{\bs C}}
\newcommand{\bG}{{\bs G}}
\newcommand{\bCvD}{{\bs C}_{v,\omega}}
\newcommand{\bK}{{\bs K}}
\newcommand{\bT}{{\bs T}}

\newcommand{\bR}{\boldsymbol{R}}

\newcommand{\bRkkD}{{\bs R}_{\kappa\kappa,\omega}}
\newcommand{\bRxx}{{\bs R}_{xx}}

\newcommand{\balpha}{\boldsymbol{\alpha}}
\newcommand{\bkappa}{\boldsymbol{\kappa}}
\newcommand{\bomega}{\boldsymbol{\omega}}

\def\R{\ensuremath{\mathrm{I\!R}}}

\newcommand*{\Scale}[2][4]{\scalebox{#1}{$#2$}}%

\newcommand{\ssum}[2]{\Scale[1.01]{\sum\limits_{\mathsmaller{#1}}^{\mathsmaller{#2}}}}

\title{Convergence analysis of kernel LMS algorithm \\ with \cred{pre-tuned} dictionary}
\name{Jie Chen$\,^{\star}$ \quad Wei Gao$\,^{\star\dagger}$ \quad C\'edric Richard$\,^{\star}$ \quad Jose-Carlos M. Bermudez$\,^{\ddagger}$}
\address{$^{\star}$ Universit\'e de Nice Sophia-Antipolis, CNRS, France\\
		$^{\dagger}$ College of Marine Engineering, Northwestern Polytechnical University, Xian, China \\
                  $^{\ddagger}$ Federal University of Santa Catarina, Florian\'opolis, SC, Brazil \\
                  \{jie.chen, cedric.richard\}@unice.fr \qquad gao\_wei@mail.nwpu.edu.cn \qquad j.bermudez@ieee.org }
\begin{document}
\ninept
\maketitle
\begin{abstract}
The kernel least-mean-square (KLMS) algorithm is an appealing tool for online identification of nonlinear systems due to its simplicity and robustness. In addition to choosing a reproducing kernel and setting filter parameters, designing a KLMS adaptive filter requires to select a so-called dictionary in order to get a finite-order model. This dictionary has a significant impact on performance, and requires careful consideration. Theoretical analysis of KLMS as a function of  dictionary setting has rarely, if ever, been addressed in the literature. In an analysis previously published by the authors, the dictionary elements were assumed to be governed by the same probability density function \cred{of the} input data. In this paper, we modify this study by considering the dictionary as part of the filter parameters to be set. This theoretical analysis paves the way for future investigations on KLMS dictionary design.
\end{abstract}
\begin{keywords}
	Kernel least-mean-square algorithm, convergence analysis, nonlinear adaptive filtering, dictionary learning
\end{keywords}
\balance
\section{Introduction}
Complex real-world applications often require nonlinear signal processing. During the last decade, adaptive filtering in reproducing kernel Hilbert spaces (RKHS) has become an appealing tool for online system identification and time series prediction~\cite{liu2011book}. By replacing inner products with a reproducing kernel, these algorithms provide an efficient and elegant way to map the input data into a high, even infinite, dimensional space with an implicit nonlinear application. The kernel recursive least-squares (KRLS) algorithm was introduced in \cite{Engel2004}. The sliding-window KRLS and the extended KRLS algorithms were derived in \cite{Vaerenbergh2006,Liu2009a}, respectively.  The kernel affine projection algorithm (KAPA) and, as a particular case, the kernel normalised LMS algorithm (KNLMS), were independently introduced in \cite{honeine2007line,Richard2009,Slavakis2008,Liu2008b}. The kernel least-mean-square algorithm (KLMS), proposed in \cite{richard2005filtrage,Liu2008a}, has attracted much attention in recent years because of its simplicity and robustness. Along the same line, the quantized KLMS algorithm (QKLMS) was recently proposed in \cite{Chen2012a}.

Kernel-based adaptive filters use more or less sophisticated criteria to construct a so-called dictionary, in order to operate with a finite-order model. This dictionary has a significant impact on performance, and requires careful consideration. One of the most informative criteria uses approximate linear dependency (ALD) condition to test the ability of the dictionary elements to linearly approximate the current  kernelized input sample \cite{Engel2004}. Other well-known criteria include the novelty criterion \cite{Platt1991}, the coherence criterion~\cite{Richard2009}, the surprise criterion \cite{Liu2009b}, and the closed-ball sparsification criterion~\cite{Slavakis2013}. Recently, KLMS algorithm with $\ell_1$-norm regularization has also been studied so that the algorithm can remove obsolete dictionary elements and then adapt to non-stationary environments~\cite{Chen2012b,Yukawa2012,gao2013kernel,gao2013l1KLMS}.

Few theoretical studies have investigated the convergence behavior of kernel-based adaptive filters compared to the number of algorithm variants that have recently been derived. This situation is partly due to technical difficulties stemming from filter nonlinearity with respect to input samples. An extended analysis of the stochastic behavior of the KLMS algorithm with Gaussian kernel was proposed in \cite{Parreira2012}, and a closed-form condition for convergence was introduced in~\cite{richard2012closed}. Stability in the mean of this algorithm with $\ell_1$-norm regularization was studied in~\cite{gao2013kernel,gao2013l1KLMS}. The aim of these works was to propose a procedure for designing KLMS filters, with appropriate parameter setting, given some \cred{desired performance}. It was assumed that the dictionary elements are governed by the same probability density function as \cred{the} input data. This situation typically arises with kernel-based adaptive filters that use a short-time sliding dictionary~\cite{richard2005filtrage,Kivinen2004,Vaerenbergh2007}. Nevertheless, this framework does not allow the \cred{user to pretune the dictionary to improve performance}. In this paper, we propose a theoretical analysis of the KLMS algorithm with Gaussian kernel that considers the dictionary as part of the filter parameters to be set. We derive models for \cred{the mean and mean-square behavior of the adaptive weights, and for the mean-square estimation error. We also determine stability conditions}. Finally, we illustrate the validity of these models with simulations. These derivations pave the way for future investigations on KLMS dictionary design.

\section{KLMS algorithm}
\subsection{Problem formulation}

Let $\cp{X}$ be a compact domain in Euclidean space $\R^L$ and let $\cp{Y} = \R$. Let $\rho_{\cp{Z}}$ be a Borel probability measure on $\cp{Z}=\cp{X}\times\cp{Y}$ whose regularity properties will be assumed as needed. Considering a sample
\begin{equation}
         \cb{z} \in \cp{Z}^N,  \qquad \cb{z} = \{(\bx(i), y(i))\}_{i=1}^{N},
\end{equation}
we aim to find a regression function $\psi$ using $\cb{z}$.  Let $\cp{H}$ be a reproducing kernel Hilbert space with kernel $\kappa: \cp{X}\times\cp{X} \rightarrow\R$. Restricting the solution to this space, the function $\psi^*$ that minimizes the regularized mean-square error
\begin{equation}
         \label{eq:regcost}
         \min_{\psi\in\cp{H}} \ssum{i=1}{N} \,  |y(i) - \psi(\bx(i))|^2 + \lambda \|\psi\|^2_{\cp{H}}, \quad\lambda \geq 0
\end{equation}
is of the form
\begin{equation}
           \label{eq:representer}
            \psi^* = \ssum{i=1}{N}\,\alpha_i \,\kappa(\cdot,\bx(i))
\end{equation}
with $\balpha=[\alpha_1, \dots, \alpha_N]^\top$ the unique solution of the well-posed linear system in $\R^N$:
\begin{equation}
            (\bK + \lambda \, \bI) \, \balpha = \by
\end{equation}
where $\bK$ is the $N \times N$ Gram matrix with $(i,j)$-th entry $\kappa(\bx(i), \bx(j))$, $\by$ is the $N \times 1$ vector with $i$-th entry $y(i)$, and $\bI$ is the $N\times N$ identity matrix. This framework is inappropriate to solve problem \eqref{eq:regcost} in an online manner, since the algorithm would suffer from the ongoing increase of the size $N$ of the model as new data arrives. A well-known strategy is to use the fixed-size model:
\begin{equation}
            \label{eq:fundic}
           \psi(\cdot) = \ssum{i=1}{M} \;\alpha_i\;\kappa(\cdot,\bxw{i}).
\end{equation}
The set $\bomega = \{\kappa(\cdot, \, \bxw{i})\}$ with $i=\{1,\dots, M\}$ is the so-called dictionary. Several rules have been proposed in the literature to construct the dictionary in an online manner. Consider a dictionary $\bomega$, and a time sequence $\{\bx(n),y(n)\}_n$, the KLMS algorithm is an efficient strategy to estimate~\eqref{eq:fundic}. The update rule is given by
\begin{equation}
	\label{eq:weightKLMS}
	\balpha(n+1) = \balpha(n) + \eta\,e(n)\,\bkappa_\omega(n)
\end{equation}
where $\bkappa_\omega(n)=[\kappa(\bx(n),\bxw{1}), \dots, \kappa(\bx(n),\bxw{M})]^\top$, and $e(n)$ is the estimation error at instant $n$
\begin{equation}
           \begin{split}
           e(n) &= y(n) - \psi(\bx(n)) \\
                   &= y(n)-\ssum{j=1}{M}\, \alpha_j(n) \,\kappa(\bx(n), \bxw{j})
           \end{split}
\end{equation}
Consider the mean-square error criterion
\begin{equation} 
      E\{e^2(n)\}  = \int_{\Omega}  \int_{\cp{X}\times\cp{Y}} e^2(n) \; d \rho_{\cp{Z}}(\bx,\by|\cp{\bomega}) \; d\rho_{\Omega}
\end{equation}
with $\rho_\Omega$ a Borel probability measure on the dictionary space $\Omega$. Except with simplified assumptions such as in~\cite{Parreira2012}, where the authors consider that the dictionary elements are governed by the same probability density function as input data, distributions of $\bomega$ generated by dictionary learning methods cannot be expressed in closed forms. In this paper, we consider the dictionary as part of the filter parameters to be set. Our objective is to characterize both transient and steady-state of the mean-square criterion conditionally to $\bomega$, that is,
\begin{equation} 
	E_Z\{e^2(n)|\bomega\}   =  \int_{\cp{X}\times\cp{Y}} e^2(n) \; d \rho(\bx,\by|\cp{\bomega}).
\end{equation}
We shall use the subscript $\omega$ for quantities conditioned on the dictionary, and $X$ to expectation with respect to input data \cred{distribution}.

\subsection{Optimal solution}
Given $\bomega$, the estimation error at instant $n$ writes
\begin{equation}
            \label{eq:eD}
            e_{\omega}(n) = y(n) - \psi_\omega(\bx(n))
\end{equation}
with $\psi_\omega(\bx(n)) = \psi(\bx(n))|\bomega$. Squaring both sides of equation~\eqref{eq:eD}, and taking the expected value, leads to the mean-square error (MSE) criterion
\begin{equation} 
            \begin{split}
             J_{\text{MSE},\omega} &= E\{e_\omega^2(n)\}  \\
                                                   &= E\{y^2(n)\} - 2\,\bp_{\kappa y,\omega}^\top \balpha + \balpha^\top\bRkkD\balpha
             \end{split} 
\end{equation}
where 
\begin{equation*}
          \bRkkD = E\{\bkappa_\omega(n)\bkappa^\top_\omega(n)|\bomega\}
\end{equation*}        
is the correlation matrix of the kernelized input, and
\begin{equation*}
           \bp_{\kappa y,\omega}= E\{y(n)\bkappa_\omega(n)|\bomega\}
 \end{equation*}
is the cross-correlation vector between $\bkappa_\omega(n)$ and $y(n)$. As $\bRkkD $ is positive definite, the optimum weight vector is given by 
\begin{equation}
	\label{eq:alphaStar}
         \begin{split}
         \balpha_{\omega}^* = \arg\min_{\balpha} J_{\text{MSE},\omega}(\balpha)  = \bRkkD^{-1}\,\bp_{\kappa y,\omega}
          \end{split}
\end{equation}
and the minimum MSE is 
\begin{equation}
            \label{eq:JminD}
         J_{\min,\omega} = E\{y^2(n)\} - \bp_{\kappa y,\omega}^\top\, \bRkkD^{-1} \, \bp_{\kappa y,\omega}.
\end{equation}

\section{Performance analysis}
We shall now derive the convergence model and stability conditions for the KLMS algorithm with Gaussian kernel, given $\bomega$. The Gaussian kernel is defined as
\begin{equation}
	\kappa(\bx_{i},\bx_{j})=\exp(-\mathsmaller{\frac{1}{2\sigma^2}}\|\bx_{i}-\bx_{j}\|^2)
\end{equation}
where $\sigma$ denotes the kernel bandwidth. Inputs $\bx(n)$ are assumed independent zero-mean Gaussian random vectors with autocorrelation matrix $\bRxx=E\{\bx(n)\bx^\top(n)\}$. Let $\bvD(n)$ be the weight error vector defined as
\begin{equation}
            \bvD(n) = \balpha(n) - \balpha_{\omega}^*.
\end{equation}

\subsection{Preliminaries and assumptions}

Before starting to derive the model, let us recall the following result on the moment generating function of any quadratic form of a Gaussian vector.  Let $\cb{\xi}=(\xi_1,\dots, \xi_L)^\top$ be a random vector following Gaussian distribution with mean and covariance matrix
\begin{equation}
            E\{\cb{\xi}\} = 0  \; \text{ and } \; \bR_{\xi\xi} = E\{\cb{\xi}\, \cb{\xi}^\top\}
\end{equation}
Let the random variable $\zeta$ be the quadratic form of $\cb{\xi}$ defined as
\begin{equation}
        \label{eq:quadvrb}
        \zeta = \cb{\xi}^\top \cb{H} \cb{\xi} + \cb{b}^\top\cb{\xi}
\end{equation}
The moment generating function of $\zeta$ is given by~\cite{Omura1965}
\begin{equation}
         \begin{split}
        \Psi_\zeta(s) = 		&|\bI - 2s\cb{H}\bR_{\xi\xi}|^{-\frac{1}{2}} \\
                                   	&\cdot \exp\left(\mathsmaller{\frac{s^2}{2}}\cb{b}^\top\bR_{\xi\xi}(\bI-2s\cb{H}\bR_{\xi\xi} )^{-1} \cb{b})\right)
        \end{split}
\end{equation}
This result will be useful to determine expected values. Simplifying assumptions are required in order to make the study of the stochastic behavior of $\bvD(n)$ mathematically feasible. The following statistical assumption is required in the analysis:

{\assumption $\bkappa_\omega(n)\bkappa_\omega^\top(n)$ is independent of $\bvD(n)$.} \\

\noindent This assumption, called modified independence assumption (MIA), is justified in detail in \cite{Minkoff2001}. It has been successfully employed in several adaptive filter analyses, and has been shown in \cite{Minkoff2001} to be less restrictive than the well known independence assumption (IA)~\cite{sayed2008adaptive}. It is called here for further reference \emph{conditioned} MIA, or CMIA, \cred{to distinguish it from the MIA used in \cite{Parreira2012}.} 
 
\subsection{Mean \cred{weight} error  analysis}

The estimation error $e_\omega(n)$ can be expressed using $\bvD(n)$ by
\begin{equation}
           e_\omega(n)= y(n) - \bkappa_\omega^\top(n)\,\bvD(n)-\bkappa_\omega^\top(n)\,\balpha^*_\omega
\end{equation}
Replacing this expression into~\eqref{eq:weightKLMS} and using the definition of $\bvD(n)$, we obtain the following recursive expression for $\bvD(n)$:
\begin{equation}
       \label{eq:recvn}
       \begin{split}
      \bvD(n+1) & = \bvD(n) + \eta\,y(n)\bkappa_\omega(n)  \\
                            & - \eta\,\bkappa_\omega^\top(n)\bv(n)\bkappa_\omega(n)  -\eta\,\bkappa_\omega^\top(n)\balpha^*_{\omega}\bkappa_\omega(n)
        \end{split}
\end{equation} 
Taking expected values of both sides of~\eqref{eq:recvn}, \cred{using CMIA and \eqref{eq:alphaStar} we have the mean weight error model}
\begin{equation}
         E\{\bvD(n+1)\} = (\bI - \eta\bRkkD)\, E\{\bvD(n)\}
\end{equation}
With the Gaussian kernel, the components of $\bRkkD $ are given by
\begin{equation}
         \label{eq:Rdd1}
         \begin{split}
         &[\bRkkD]_{ij} \\ 
         	& = E_{X}\!\left\{\exp\!\left(-\mathsmaller{\frac{1}{2\sigma^2}}\left[\|\bx(n)-\bxw{i}\|^2+\|\bx(n)-\bxw{j}\|^2\right]\right)\right\} \\
		& = \exp\!\left(-\mathsmaller{\frac{1}{2\sigma^2}}\left[\|\bxw{i}\|^2+\|\bxw{j}\|^2\right]\right) \\ 
                 &    \hspace{+4mm}\cdot E_{X}\!\left\{\exp\!\left(-\mathsmaller{\frac{1}{\sigma^2}}\big[\|\bx(n)\|^2-(\bxw{i}+\bxw{j})^\top\bx(n)\big]\right)\right\}
         \end{split}
\end{equation}
Let $\bar{\bx}_{\omega_{ij}} = \bxw{i}+\bxw{j}$ and $\|\bar{\bx}_{\omega_{ij}}\|^{(2)} =\|\bxw{i}\|^2+\|\bxw{j}\|^2$. Comparing the second term on the RHS of~\eqref{eq:Rdd1} with~\eqref{eq:quadvrb} \cred{with $\cb{H}=\cb{I}$, $\cb{b} = -\bar{\bx}_{\omega_{ij}}$ and $s=-\frac{1}{\sigma^2}$} we get
\begin{equation}
	\label{eq:Rdd2}
 	\begin{split}
	[\bRkkD]_{ij}  	&	=\exp\!\left(-\mathsmaller{\frac{1}{2\sigma^2}}\|\bar{\bx}_{\omega_{ij}}\|^{(2)} \right)
						|\bI+\mathsmaller{\frac{2}{\sigma^2}}\bRxx|^{-\frac{1}{2}} \\
				&  	\hspace{-9mm} \cdot\exp\!\left(\mathsmaller{\frac{1}{2\sigma^4}}
						\bar{\bx}_{\omega_{ij}} ^\top{\bRxx(\bI+\mathsmaller{\frac{2}{\sigma^2}}\bRxx)^{-1}}\bar{\bx}_{\omega_{ij}}\right)
         \end{split}
\end{equation}
In order to express this formula in a more compact manner, we use the identity $(\bI+\cb{A}^{-1})^{-1} = \cb{A}(\cb{A}+\bI)^{-1}$ with ${\bRxx(\bI+\mathsmaller{\frac{2}{\sigma^2}}\bRxx)^{-1}}$. This yields
\begin{equation}
	\label{eq:Rdd3}
	\begin{split}
	[\bRkkD]_{ij} 	&= |\bI + \mathsmaller{\frac{2}{\sigma^2}}\bRxx|^{-\frac{1}{2}} \\
				&  \hspace{-7mm}\cdot\exp\!\left(-\mathsmaller{\frac{1}{4\sigma^2}}
				\left[2\|\bar{\bx}_{\omega_{ij}}\|^{(2)}-\|\bar{\bx}_{\omega_{ij}}\|^2_{(\bI+ \sigma^2\bRxx^{-1}/2)^{-1}}\right]\right)
         \end{split}
\end{equation}

{\noindent\theorem (Stability in the mean) Assume CMIA holds. Then, for any initial condition, given a dictionary $\bomega$, the Gaussian KLMS algorithm~\eqref{eq:weightKLMS} asymptotically converges in the mean if the step size is chosen to satisfy
\begin{equation}
	\label{eq:stepsize1}
	\begin{split}
		0 < \eta < \frac{2}{\lambda_{\max}(\bRkkD)}
	\end{split}
\end{equation}
where $\lambda_\text{max}(\cdot)$ denotes the maximum eigenvalue of the matrix. The entries of $\bRkkD$ are given by~\eqref{eq:Rdd3}}.

\subsection{Mean-square error analysis}
The second-order moments of the weights are related to the MSE through~\cite{Parreira2012}
\begin{equation}
           \label{eq:Jn}
          J_{\text{MSE},\omega}(n) = J_{\min,\omega} + \text{trace}\{\bRkkD\bCvD(n)\}
\end{equation}
where $\bCvD(n)$ denotes the correlation matrix of the weight error vector $\bvD(n)$, i.e.,
\begin{equation}
          \bCvD(n) = E\{\bvD(n)\,\bvD^\top(n)\},
\end{equation}
and $J_{\min,\omega}$ is the minimum MSE given by~\eqref{eq:JminD}. The second term on the RHS of~\eqref{eq:Jn} is the excess mean-square error (EMSE), denoted by $J_{\text{EMSE},\omega}(n)$. Estimating $J_{\text{MSE},\omega}(n)$ requires a model for $\bCvD(n)$. 

Post-multiplying \eqref{eq:recvn} by its transpose, taking the expectation, and using CMIA, we obtain the following recursion
\begin{equation}	
      \label{eq:Cv}
      \begin{split}
              \bC_{v,\omega}(n+1) \approx\;&  \bC_{v,\omega}(n)  + \eta^2\,\T_\omega + \eta^{2}\,\bRkkD J_{\min,\omega}  \\
                                                  &- \eta\,(\bRkkD \bC_{v,\omega}(n) + \bC_{v,\omega}(n)\bRkkD)
       \end{split}
\end{equation}
with 	
\begin{equation}
 	\label{eq:T}
  	\T_{\omega} = E\{\bkappa_\omega(n)\,\bkappa_\omega^{\top}(n)\,\bvD(n)\,\bvD^{\top}(n)\,\bkappa_\omega(n)\,\bkappa_\omega^{\top}(n)\}.
\end{equation}
Evaluating~\eqref{eq:T} is a challenging step in the analysis. In~\cite{Parreira2012}, for independent Gaussian-distributed dictionary elements, this leads to extensive calculations of up to eighth-order moments of $\bx(n)$. In~\cite{gao2013l1KLMS} we provided a greatly simplified alternative to this. However both  situations do not match the framework developed this paper, since dictionary elements are now considered as known. In order to determine the expected value of $\bT_\omega$, assuming CMIA holds, the $(i,j)$-th entry of $\bT_\omega$ writes:
\begin{equation}
	\label{eq:expT}
	\begin{split}
		[\bT_\omega]_{ij} \approx\ssum{\ell=1}{M} \ssum{p=1}{M} &E_X\{\kappa_{\omega,i}(n)\,\kappa_{\omega,j}(n)
		\kappa_{\omega,\ell}(n)\,\kappa_{\omega,p}(n)  \}  \\
		&\cdot [\bC_{v, \omega}(n)]_{\ell p}.     	
	\end{split}
\end{equation}
 where $\kappa_{\omega,i}(n)=\kappa(\bx(n),\bx_{\omega_i})$. Let us define the matrix $\bK_{\omega}^{(i,j)}$ with $(\ell,p)$-th entry 
 \begin{equation}
	[\bK_{\omega}^{(i,j)}]_{\ell p} = E_X\{\kappa_{\omega,i}(n)\kappa_{\omega,j}(n)\kappa_{\omega,\ell}(n)\kappa_{\omega,p}(n)\}.
 \end{equation}
Expression~\eqref{eq:expT} can be rewritten as
\begin{equation}
	\label{eq:expTK}
	[\T_{\omega}(n)]_{ij} \approx \tr \{\bK_{\omega}^{(i,j)}\bC_{v,\omega}(n) \}
\end{equation}
In order to determine~\eqref{eq:expTK}, we need to evaluate the expected values in $[\bK_{\omega}^{(i,j)}]_{\ell p}$. Let us introduce further the notations 
\begin{equation}
	\begin{split}
		&\bar{\bx}_{\omega_{ij\ell p}} 			= \bxw{i}+\bxw{j}+\bxw{\ell}+\bxw{p} \\
		&\|\bar{\bx}_{\omega_{ij\ell p}}\|^{(2)} 	= \|\bxw{i}\|^2+\|\bxw{j}\|^2+\|\bxw{\ell}\|^2+\|\bxw{p}\|^2
	\end{split}
\end{equation}
We have
\begin{equation}
        \label{eq:Kdij}
        \begin{split}
        [\bK_{\omega}^{(i,j)}&]_{\ell p} \\   &= E_X\{\kappa_{\omega,i}(n)\,\kappa_{\omega,j}(n)\,\kappa_{\omega,\ell}(n)\,\kappa_{\omega,p}(n)\}  \\
        &= E_X\Big\{\!\exp\!\Big(\mathsmaller{-\frac{1}{2\sigma^2}}  \ssum{k=\{i,j,\ell,p\}}{}\|\bx(n)-\bx_{\omega_k}\|^2 \Big)\Big\}\\
        &= \exp\!\left(-\mathsmaller{\frac{1}{2\sigma^2}}\|\bar{\bx}_{\omega_{ij\ell p}}\|^{(2)}\right) \\
            &\hspace{+4mm}\cdot E_{X}\!\left\{\exp\!\left(-\mathsmaller{\frac{1}{\sigma^2}}[2\|\bx(n)\|^2- \bar{\bx}_{\omega_{ij\ell p}} ^\top\, \bx(n)]\right)\right\}
        \end{split}
 \end{equation}
Now setting $\cb{H} = 2\bI$, $\cb{b} = -\bar{\bx}_{\omega_{ij\ell p}}$ and $s=-\frac{1}{\sigma^2}$ in \eqref{eq:quadvrb}, we get
\begin{equation}
	\label{eq:Kdij}
	\begin{split}
	[\bK_{\omega}^{(i,j)}]_{\ell p}	
		&= |\bI+\mathsmaller{\frac{4}{\sigma^2}}\bRxx|^{-\frac{1}{2}} \\
		&  	\hspace{-11mm}\cdot\exp\!\left(-\mathsmaller{\frac{1}{8\sigma^2}}
			\left[4\|\bar{\bx}_{\omega_{ij\ell p}}\|^{(2)}-\|\bar{\bx}_{\omega_{ij\ell p}}\|^2_{(\bI+ \sigma^2\bRxx^{-1}/4)^{-1}}\right]\right)
         \end{split}
\end{equation}

\begin{figure*}[!t]
   	\includegraphics[trim = 30mm 15mm 30mm 20mm, clip, scale=0.55]{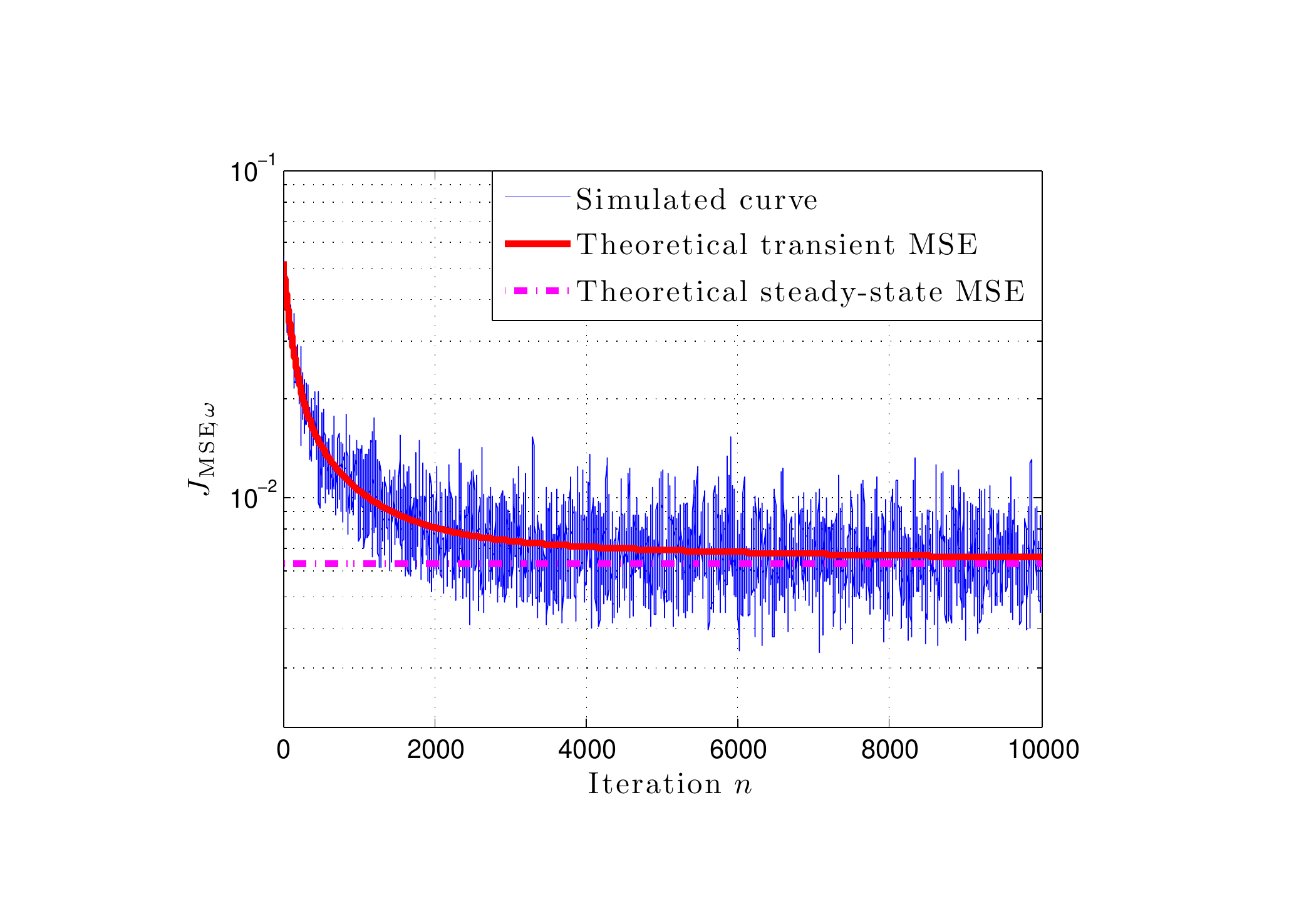}  \qquad
	 \includegraphics[trim = 30mm 15mm 30mm 20mm, clip, scale=0.56]{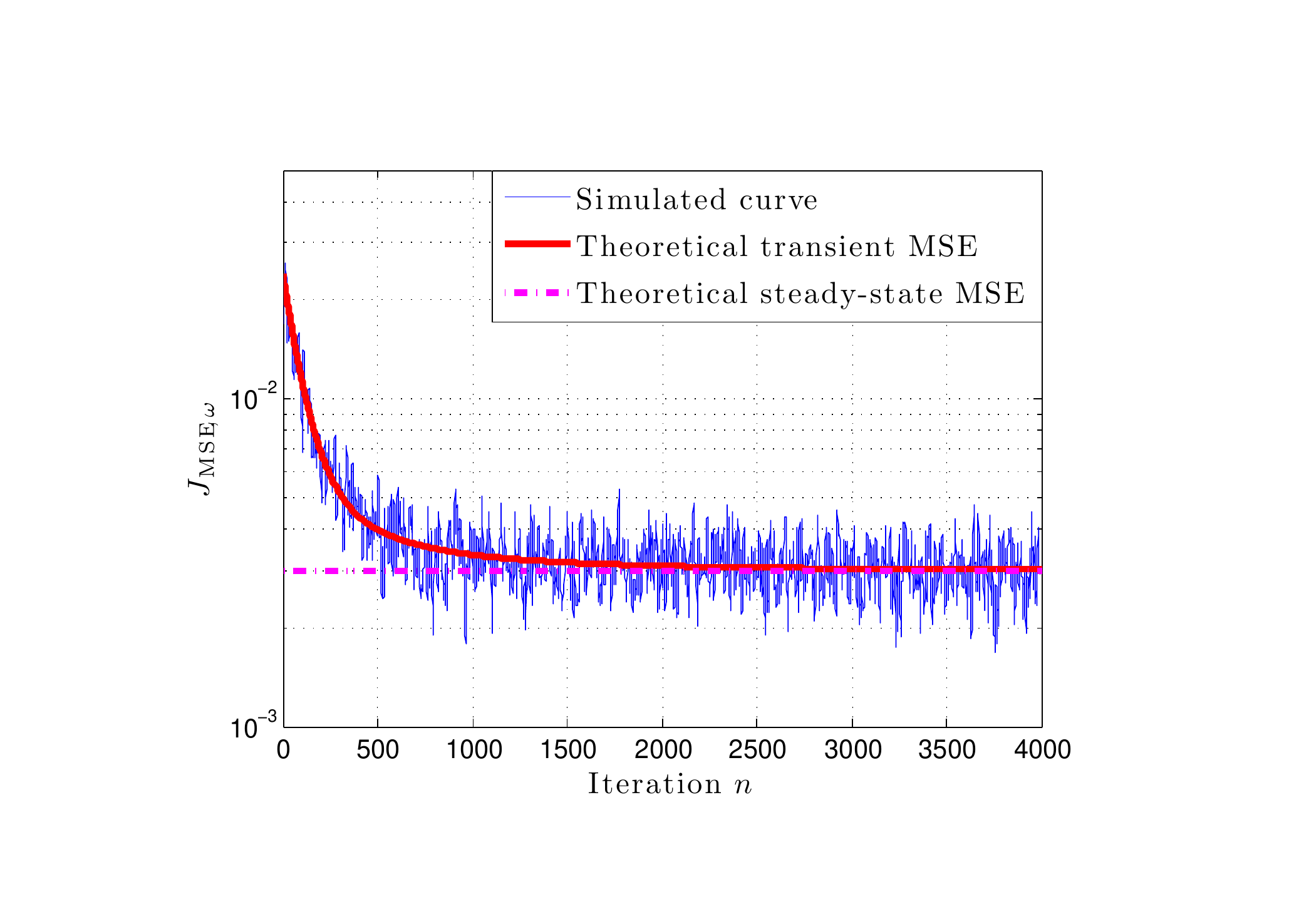} 
	 \vspace{-4mm}
	\caption{Simulation results (Left: learning curves for system 1. Right: Learning curves for system 2).}
	\label{fig:simu}
\end{figure*}

\noindent With this expression, recursion \eqref{eq:Cv} can readily be evaluated. In lexicographic form, i.e., the columns of
each matrix are stacked on top of each other into a vector, equation~\eqref{eq:Cv} becomes
\begin{equation}
	\label{eq:lex.C}
	\bc_{v,\omega}(n+1) = \bG_\omega\,\bc_{v,\omega}(n) + \eta^2 J_{\text{min},\omega}\,{\cb{r}}_{\kappa\kappa,\omega}
\end{equation}
with
\begin{equation}
          \label{eq:vecCRkk}
           \bG_\omega = \bI - \eta(\bG_{\omega,1}+ \bG_{\omega,2})+ \eta^2\bG_{\omega,3}
\end{equation}
where $\bc_{v,\omega}(n)$ and ${\cb{r}}_{\kappa\kappa,\omega}$ are the lexicographic representations of matrices $\bC_{v,\omega}(n)$ and $\bRkkD$, respectively. Matrix $\mG_\omega$ is given by:  
\begin{itemize}
          \item $\bI$ is the identity matrix of dimension $M^2 \times M^2$; 
          \item $\bG_{\omega,1}=\bI \otimes \bRkkD$, where $\otimes$ denotes the Kronecker product; 
          \item $\bG_{\omega,2}=\bRkkD \otimes \bI$; 
          \item $\bG_{\omega,3}$ entries are: $[{\bG_{\omega,3}}]_{i+(j-1)M,\ell+(p-1)M} =  [\bK_{\omega}^{(i,j)}]_{\ell,p}  $ with $1 \leq i,j,\ell,p \leq M$.
\end{itemize}
Note that $\mG_{\omega,1}$ to $\mG_{\omega,3}$ are symmetric matrices, which implies that the matrix $\mG_\omega$ is also symmetric. The following results directly come from \eqref{eq:lex.C}--\eqref{eq:vecCRkk}:
{\theorem (Mean-square stability) Assume CMIA holds. For any initial condition, given a dictionary $\bomega$, the Gaussian KLMS algorithm~\eqref{eq:weightKLMS} is mean-square stable if the matrix $\bG_\omega$ is stable.
}

{\theorem (Mean-squared error) Consider a sufficiently small step size $\eta$, which ensures mean and mean-square stability. The steady-state MSE is  given by \eqref{eq:Jn} with the lexicographic representation of $\bC_{v,\omega}(\infty)$ given by
\begin{equation}
	\label{eq:c_inf}
	\vc_{v,\omega}(\infty) = \eta^2\,J_{\textnormal{min},\omega} \,(\mI - \mG_\omega)^{-1}\,{\bs r}_{\kappa\kappa,\omega}.
\end{equation}
}

\vspace{-5mm}

\section{Experiment}
In this section, we consider two problems of nonlinear system identification with KLMS. We shall compare simulated learning curves and analytical models to validate our approach.  

In the first experiment, we considered the input sequence
\begin{equation}
         \label{eq:iirx}
         x(n) =  \rho\, x(n-1) + \sigma_x\,\sqrt{1-\rho^2} \,w(n)
\end{equation}
with $w(n)$ a noise following the i.i.d. standard normal distribution. The nonlinear system was defined as follows:
\begin{equation}
    \left\{
\begin{split}
        & u(n) = 0.5\,x(n) -0.3\,x(n-1) \\
        & y(n) = u(n) - 0.5\,u^2(n)+0.1\, u^3(n) + v(n)
\end{split} \right.
\end{equation}
where $v(n)$ is an additive zero-mean Gaussian noise with standard deviation $\sigma_{v}=0.05$. At each instant, the KLMS algorithm was updated based on the input vector $\bx(n) = [x(n), x(n-1)]^\top$ and the reference signal $y(n)$. We set $\sigma_x=0.5$ and $\rho = 0.5$. The Gaussian kernel with bandwidth $\sigma=0.25$ was used.  Twenty-five samples on a uniform grid defined on $[-1, 1]\times[-1, 1]$ were \cred{randomly} selected to be the dictionary elements $\bxw{i}$, $i=1,\dots, 25$. The step size was set to $\eta = 0.05$. The learning curves of the algorithm are depicted in Fig.~\ref{fig:simu} (left). The simulated curves were obtained by averaging over 100 Monte-Carlo runs. Theoretical MSE evolution was estimated by~\eqref{eq:Jn}, and $\bC_{v,\omega}(n)$ was recursively evaluated with expression \eqref{eq:lex.C}. The steady-state MSE was calculated by Theorem 3. It can be noticed that although inputs $\bx(n)$ are correlated in time, theoretical results derived with CMIA accurately describe the behavior of the KLMS algorithm.

In the second experiment, we considered the fluid-flow control problem studied in~\cite{Duwaish1996,Voros2003}. The input signal was also generated with~\eqref{eq:iirx} with $\sigma_x=0.25$ and $\rho=0.5$. The nonlinear system was defined by
\begin{equation}
    \left\{
\begin{split}
          & u(n) = 0.1044\,x(n) +0.0883\,x(n-1) \\
           &\hspace{1.5cm}            +1.4138\,y(n-1)-0.6065\,y(n-2) \\
          & y(n) = 0.3163\,u(n)/\sqrt{0.10+0.90\, u^2(n)} + v(n)
     \end{split} \right.    
\end{equation}
where $v(n)$ is an additive zero-mean Gaussian noise with standard deviation $\sigma_{v}=0.05$. A Gaussian kernel with bandwidth $\sigma=0.15$ was used. Thirty-seven dictionary elements were pre-selected with the \emph{coherence criterion}~\cite{Richard2009}. The step size was set to $\eta = 0.05$. We ran the KLMS algorithm over $100$ Monte-Carlo to empirically estimate its performance, and we evaluated the theoretical model. This led us to the learning curves presented in Fig.~\ref{fig:simu} (right). This simulation also confirms the validity of our theoretical analysis.

\section{Conclusion and perspectives}

Designing a KLMS-type adaptive filter requires to select a dictionary, which has a significant impact on performance and thus requires careful consideration. In this paper, we derived a theoretical model to characterize the convergence behavior of the KLMS algorithm with Gaussian kernel. This model depends on dictionary setting, which can now be considered as part of the filter parameters to be set. In future work, we will exploit this additional flexibility to design  application-dependent dictionaries.

\newpage

\balance
\bibliographystyle{IEEEbib}
\bibliography{ref}

\end{document}